\documentclass{nle}

\usepackage{apalike}
\let\bibhang\relax
\usepackage{natbib}

\usepackage[english]{babel}
\usepackage[utf8]{inputenc}
\usepackage[T1]{fontenc}

\usepackage{comment}
\usepackage{url}
\usepackage{amsmath}
\usepackage{bm}
\usepackage{xspace}
\newcommand\BLEU{\textsc{Bleu}\xspace}
\usepackage{fp}
\def\roundposition{1}
\edef\rounded{0}
\newcommand{\rdm}[1]{\edef\rounded{0}\FPeval\rounded{round(#1,\roundposition)}\rounded}

\usepackage{textcomp}
\newcommand{\ftextnumero}{{\fontfamily{txr}\selectfont \textnumero}}

\title[]{Modeling Target-Side Morphology \\in Neural Machine Translation: \\A Comparison of Strategies}
\author[Weller-Di Marco, Huck, Fraser]
       {Marion Weller-Di Marco, Matthias Huck, Alexander Fraser
         \thanks{The first two authors had equal effort.
           The first author has conducted part of the research for this work while being at ILLC, University of Amsterdam.
           The second author has conducted the research for this work while employed at CIS, LMU Munich.
         }\\
         Center for Information and Language Processing, LMU Munich}

\pagerange{\pageref{firstpage}--\pageref{lastpage}}
\pubyear{2019}

\usepackage{color}

\begin{document}

\label{firstpage}
\maketitle

\begin{abstract}
Morphologically rich languages pose difficulties to machine translation.
Machine translation engines that rely on statistical learning from parallel
training data, such as state-of-the-art neural systems, face challenges
especially with rich morphology on the output language side. 
Key challenges of rich target-side morphology in data-driven machine
translation include:
{\small (1)} A large amount of differently inflected word surface forms
  entails a larger vocabulary and thus data sparsity, which in turn
  aggravates the learning problem. 
{\small (2)} Some inflected forms of infrequent terms typically do not appear
  in the training corpus, especially under low-resource conditions, which
  makes closed-vocabulary systems unable to generate these unobserved
  variants.
{\small (3)} Linguistic agreement requires the system to correctly match the
  grammatical categories between inflected word forms in the output sentence,
  both in terms of target-side morphosyntactic wellformedness and semantic
  adequacy with respect to the input.

These challenges can be tackled with dedicated linguistic modeling of
morphology on the target language side. 
Recent research has shown that such modeling considerably improves the
machine translation quality of neural systems that are based on shallow-RNN
encoder-decoder architectures with attention \citep{bahdanau+al-2014-nmt}. 
In this paper, we re-investigate two target-side linguistic processing
techniques---a \emph{lemma-tag} strategy \citep{tamchyna17:wmt} and a \emph{linguistically informed
word segmentation} strategy \citep{huck17:wmt}---with one of the latest incarnations of neural
architectures for machine translation, namely the \emph{Transformer} model \citep{vaswani2017attention}.

Our series of empirical experiments are conducted on a English$\to$German
translation task under three training corpus conditions of different
magnitudes.
We find that the stronger Transformer baseline leaves less room for improvement
than the older shallow-RNN encoder-decoder model when translating in-domain.
However, we find that linguistic modeling of target-side morphology does
benefit the Transformer model when the same system
is applied to out-of-domain input text. We also successfully apply our approach
to English to Czech translation.

\end{abstract}

\pagebreak

\section{Introduction}
Neural machine translation (NMT) has become state-of-the-art in machine translation in the last years. 
It has been shown to surpass phrase-based statistical machine translation both with regard
to general performance in terms of automatic metrics such as \BLEU \citep{papineni-EtAl:2002:ACL}, and also in 
terms of handling linguistic aspects such as syntax and morphology,
see, e.g., \citet{bentivogli-EtAl:2016:EMNLP2016}.
However, despite the increase in performance, standard NMT systems do not make strong use of morphological
information, but rather benefit from better access to
contextual information in the source and target sentences than was possible in
phrase-based statistical machine translation.

Recent research has shown that the integration of linguistic information can 
further improve the translation quality, with a  main focus on the syntactic 
and morphological levels. For example, \citet{eriguchi:acl2016} and \citet{bastings:emnlp2017}
demonstrated the positive impact of integrating source-side syntactic information; 
\citet{nadejde-etal-2017-predicting} showed that using syntactic information on the source 
and target side of an NMT system leads to improved translation quality. 
Addressing the morphological level, \citet{tamchyna17:wmt}  and \citet{huck17:wmt}
presented systems to generate target-side inflected forms, resulting in improved translation
quality for the respective settings.

A major morphological
problem in many applications of natural language processing is the lack of
generalization when only looking at surface forms as they appear in a text:
inflectional variants of the same lemma are not recognized as closely related, but are
treated as completely different words, which is obviously ineffective.
In particular for morphologically complex languages with many surface forms, the
richness of surface forms increases the vocabulary size, and subsequently leads to data 
sparsity problems. 
Furthermore, when translating into a language with complex inflectional morphology,
there is the problem of selecting the correct inflection from the set of all possible
inflections that may or may not have been observed in the training data.
In addition to inflection, there is the issue of productive word formation, 
i.e., the creation of potentially new complex words from observed components.
To address the range of problems that comes along with rich target-side morphology, 
a strategy that 
\begin{itemize}
  \item introduces a more general, effective and linguistically sound representation, 
  addressing the issue of generalization, and 
  \item allows for the generation of new forms, ideally according to linguistic principles
\end{itemize}
constitutes a promising and powerful tool to handle translation into morphologically
complex languages.

In this paper, we present both a knowledge-rich and a knowledge-poor approach to handle
target-side morphology, and explain the linguistic intuitions and pros and cons of the 
two approaches versus a linguistically uninformed method (Byte Pair Encoding) with respect to
generalization.
We additionally highlight the important issue of the creation of novel surface words 
(words which were not seen in the training data), showing how and where this can occur 
in both the knowledge-rich and the knowledge-poor approaches.
Our analysis shows that we can achieve linguistic generalization through the use of 
the two presented knowledge-rich and  knowledge-poor approaches, and that this outcome 
is particularly  strong in out-of-domain translation scenarios.

\subsection{Outline}

We give an overview of the problem in Section~\ref{sec:overview},
briefly introducing the approaches compared in this paper, and then
describe
our two strategies to model target-side morphology in detail -- the
generation of inflected forms from an abstract representation of
lemmas and tags in Section~\ref{sec:lemma-tag-approach} and
target-side word segmentation in
Section~\ref{sec:target-side-segmentation}.
In our experiments in Section~\ref{sec:experiments}, the lemma-tag and the
segmentation systems are first compared on the English--German language pair
for three corpus settings, namely with small, medium, and large training data,
using the {\it Transformer} NMT model. The systems are applied to in-domain test 
data (news text), and then analyzed in an out-of-domain context (medical domain). 
Next, we compare the same strategies within another neural model, a shallow-RNN
encoder-decoder translation system, since this model was used in prior work on
target-side morphology in NMT. 
Further analysis and discussion are provided in Sections~\ref{sec:discussion}
and \ref{section:translation-examples}.
We also adapt the strategies to another language pair, English--Czech in Section~\ref{sec:en-cz},
where we find that large gains are achieved when only small training data is available.
In Section~\ref{sec:rel-work}, we review related work before
concluding the paper in Section~\ref{sec:conclusion}.

\section{Overview: Target-side morphology in neural machine translation}
\label{sec:overview}

Neural machine translation (NMT) is a very successful approach to
translation, and has recently resulted in large gains in translation
quality, particularly when applied to very large training data
sets. Two prominent examples of popular models are shallow recurrent
neural network encoder-decoder architectures with attention
\citep{bahdanau+al-2014-nmt}, and the \emph{Transformer} model
\citep{vaswani2017attention}, which uses self-attention instead of
recurrence.

For NMT systems, the overall vocabulary size can be problematic -- an
unrestricted vocabulary results in memory problems and intractable
training times.  A common approach to restrict the vocabulary size in
NMT has been presented by \citet{sennrich-bpe:acl2016}. 
They adopt a technique in the manner of the Byte Pair Encoding (BPE)
compression algorithm \citep{Gage:1994:NAD:177910.177914} in order to segment
words into smaller sub-word units. 
The BPE word segmenter conceptionally proceeds by first splitting all words in
the whole corpus into individual characters. The most frequent adjacent pairs
of symbols are then consecutively merged, until a specified limit of merge
operations has been reached. Merge operations are not applied across word
boundaries. The merge operations learned on a training corpus can be stored
and applied to other data, such as test sets.

While this strategy has been shown to be effective, it is not
linguistically informed, and thus leads to non-optimal splittings. An
example of a non-optimal splitting is a splitting along non-linguistic
boundaries which blocks linguistic productivity. Furthermore, BPE does
not offer a satisfactory solution to another important problem arising when
translating {\it into} a morphologically rich language -- namely the
selection or generation of correctly inflected surface forms given the
sentence context.

Recently, linguistically motivated strategies have been proposed to address the 
problems of obtaining a manageable vocabulary size while not losing coverage
\citep{ataman:eamt2017,ataman:acl2018},
as well as handling target-side inflection for morphologically rich languages \citep{burlot-etal-2017-word,10.1007/978-3-319-68456-7_2,passban18:naacl,conforti18:amta}.
In this paper, we present a case study that contrasts two conceptually similar
strategies, of which one makes heavy use of linguistic resources such as parse
information and a tool for morphological analysis and generation \citep{tamchyna17:wmt}, 
whereas the other approach is comparatively knowledge poor, using only a stemmer \citep{huck17:wmt}.

\paragraph{Knowledge-rich approach.} \citet{tamchyna17:wmt} replace inflected word forms 
on the target side to a pair of lemma and morphological tag in the training data of NMT 
systems from the language pairs English--Czech and English--German. The output of these 
systems is re-inflected by generating inflected forms from the tag-lemma pairs in a 
deterministic post-processing step. Reducing inflected word forms to lemmas and a 
comparatively small set of tags greatly decreases the number of observed word types in 
the training data, while the morphological tags allow for meeting agreement constraints 
in the generation step.
  
\paragraph{Knowledge-poor approach.} \citet{huck17:wmt} propose a segmentation strategy 
that separates inflectional suffices from the word stems, as well as a linguistically 
sound splitting of complex stems, e.g. the handling of prefixes/suffixes and compound 
splitting. After translation, the components are simply put back together to form 
inflected target-language words. 

\paragraph{Differences.} Both strategies apply the concept of reducing inflected word forms 
to stems during the training and translation process and a post-processing step to obtain 
inflected surface forms, leading to an improvement in the respective studied translation 
tasks. There are two main differences between the two approaches: (i) the use of explicit
linguistic information (i.e. tags annotated with morphological features) and morphological
resources in \citet{tamchyna17:wmt}, in contrast to the comparatively
resource-poor strategy without explicit generation step in \citet{huck17:wmt}; and (ii)
the handling of sub-words through a more sophisticated segmentation approach by 
\citet{huck17:wmt}, which is not addressed by \citet{tamchyna17:wmt}.

The comparison between these two strategies looks at the performance of different
training data sizes for an English--German news translation task, and the application
to out-of-domain test sets (medical domain).
They are also applied to English--Czech.


\begin{table}
\caption{Translation output in lemma-tag format and the resulting inflection for the input sentence 
``{\it the EU commission wants to double the limits for mercury in large predatory fish \ldots}''}
{\small
\setlength\tabcolsep{0.2em}
\begin{tabular}{llll}
\hline
\hline
{\bf Tag}                   & {\bf Lemma}  & {\bf Inflected } & {\bf Gloss}\\
                            &              & {\bf Form}       & \\
\hline
{\tt <+ART><Fem><Nom><Sg><St>}    & {\tt die<Def>}               & die           & {\it the} \\
{\tt <+NN><Fem><Nom><Sg><NA>}     & {\tt EU-<TRUNC>Kommission}   & EU\,-\,Kom-   & {\it EU com-}\\[-1mm]
                                  &                              & mission       & {\it mission}\\
{\tt <+V><1><Sg><Pres><Ind>}      & {\tt wollen}                 & will          & {\it wants} \\
{\tt <+ART><NoGend><Acc><Pl><St>} & {\tt die<Def>}               & die           & {\it the} \\
{\tt <+NN><Fem><Acc><Pl><NA>}     & {\tt Grenze}                 & Grenzen       & {\it limits}\\ 
{\tt [APPR-Acc]}                  & {\tt f\"ur}                  & f\"ur         & {\it for}\\
{\tt <+NN><Neut><Nom><Sg><NA>}    & {\tt Quecksilber}            & Quecksilber   & {\it mercury}\\
{\tt [APPR-Dat]}                  & {\tt in}                     & in            & {\it in}\\
{\tt <+ADJ><NoGend><Dat><Pl><St>} & {\tt gro\ss <Pos>}           & gro\ss en     & {\it large}\\
{\tt <+NN><Masc><Dat><Pl><NA>}    & {\tt Raub<NN>Fisch}          & Raubfischen   & {\it predatory fish}\\
{\tt <+V><Inf>}                   & {\tt verdoppeln}             & verdoppeln    & {\it double}\\
\hline
\end{tabular}
}
\label{table:lemma-tag-example}
\end{table}


\section{Knowledge-rich approach: The lemma--tag strategy}
\label{sec:lemma-tag-approach}
The approach for generating inflected forms from pairs of lemmas and morphological tags consists of
two steps: translation into an abstract representation and generation of inflected forms. 
It essentially follows the concept presented by \citet{tamchyna17:wmt}.\footnote{The main difference to the
work by \citet{tamchyna17:wmt} is the representation of the source-side data: here, we use plain English
on the source side for a better comparability with the target-side segmentation approach, whereas a sequence 
of tag-lemma was used by \citet{tamchyna17:wmt} for the English$\to$German experiment.} 

To build the translation model, inflected surface forms in the training data, e.g. the adjective form 
{\it gr\"unes} ({\it green}), are replaced by an abstract representation of pairs of lemmas and tags 
annotated with the  respective morphological features, e.g. {\it gr\"un\,+\,ADJ-Neut.Acc.Sg.Wk}.
After translation, the system output is re-inflected in a deterministic post-processing 
step, where the lemma-tag pairs are transformed into inflected surface forms by means of a morphological 
resource (e.g. {\it\,gr\"un +\,ADJ-Neut.Dat.Sg.Wk $\rightarrow$ gr\"unem}).

As the lemma-tag pairs exactly correspond to the surface forms, there is no loss of information. 
At the same time, a translation model trained on this data representation has access to the more general 
lemma in combination with flat syntactic information in the form of the morphologically annotated tags.

\subsection{Abstract representation}
\label{sec:training-data-lemma-tag}
To prepare the training data for the lemma-tag strategy, information obtained from a parser 
is combined with the morphological tool SMOR \citep{schmid04:lrec}, a finite-state based morphological
resource to analyze and generate word forms.

SMOR is used to obtain the lemma of a surface form, such as 
{\it B\"aume$_{Inflected}$ $\rightarrow$ Baum$_{Lemma}$} ({\it trees$_{Pl}$/tree$_{Sg}$}).
As inflected word forms on their own can be ambiguous (for example, {\it B\"aume} has the possible 
case values {\it nominative}, {\it accusative} and {\it genitive}), a morphological analysis based on SMOR 
alone, i.e. without sentence context, is not sufficient. 
Thus, the German data is parsed with BitPar \citep{schmid04:coling} in order to obtain the
morphological annotation\footnote{The parse structure itself is not needed; even though 
it could make for an interesting addition, for example building on the work by \citet{aharoni:acl2017} 
who use linearized constituent trees on the source-side of an NMT system.}, which is used as basis for 
the morphological tag, while SMOR only provides the lemma of a word.

While the reduction of inflected words to lemmas decreases the vocabulary count, a further reduction by
means of BPE is still required in order to reach the desired vocabulary size.
The lemma-tag strategy only addresses inflected forms, whereas other words such as proper names, which
contribute considerably to the vocabulary count, remain unchanged. 
(Table \ref{table:vocabulary-size} gives an overview on the vocabulary size of the baseline vs. the 
lemma-tag representation.) 
Furthermore, BPE segmentation is applied to both the source and the target data in order to also reduce 
the vocabulary on the source-side. 

\subsection{Inflection}
Table~\ref{table:lemma-tag-example} illustrates the post-processing step for the generation of inflected
forms: the left side shows the tag-lemma sequences
output by the translation system, based on which inflected forms are generated using the morphological 
tool SMOR. The tags contain all relevant features for nominal inflection ({\it gender},
{\it case}, {\it number}, {\it strong/weak}) and verbal inflection ({\it person}, {\it number}, {\it tense}, 
{\it mood}), and in combination with the lemma, the inflected form is unambiguously defined.\footnote{If there 
are orthographic variants the most frequent form according to a monolingual word list is chosen.}

\subsection{Generation of new forms and word formation}
\label{sec:generation-lemma-tag}
The lemma-tag strategy addresses some of the main problems in NMT: vocabulary size and the accompanying lack
of generalization in morphologically rich languages, and the selection of context-appropriate forms. 
By representing inflected forms as a pair of lemmas and tags, the vocabulary size can be considerably reduced 
in a linguistically sound way, and the translation system can generalize over inflectional variants.
The generation step makes use of the explicit linguistic information contained in the morphologically annotated 
tags, such that the inflected forms fit into the sentence context and morpho-syntactic agreement 
constraints are met.

In particular, it is possible to generate inflectional variants not occurring in the training data. 
While this is to a certain extent also possible with BPE splitting, the lemma-tag approach enables
a systematic generation, whereas generation based on BPE segments depends on ``lucky splitting''
into lemma and suffix and ``lucky recombination'' into a valid and contextually fitting word form.
Furthermore, BPE segmentation cannot handle non-concatenative operations.

While the lemma-tag strategy does not explicitly address word formation, in some cases there are
indirect benefits coming from the lemma-internal representation, namely an analysis in terms of 
derivation and word formation, as can be seen in the example below:

\begin{itemize}
  \item [] Planetenbewegungen (`planetary motion') \\[-2ex]
  \item [] {\tt Planet<NN>bewegen<V>ung<SUFF><+NN><Fem><Gen><Pl>} 
  \item [] {\it planet$_{NN}$ ~ move$_{V}$ ~~~~~ ment$_{SUFF}$ }
\end{itemize}

\noindent
While the derivational information is not actively used in the translation system, the structure of 
the lemma representation can have an indirect benefit by normalizing components of morphologically 
complex words such as compounds. 
In the example above, for instance, the modifier {\it Planeten} is replaced by the lemma {\it Planet},
i.e. dropping the transitional element {\it -en}. 
Similarly, the verb stem {\it beweg-} is expanded to the full form {\it bewegen} ({\it to move}).
Thus, the BPE segmentation process does not have  to deal with transitional elements or other variation,
but can benefit from a consistent representation between all occurrences of a word or sub-word.
In particular, the normalization to lemmas in the modifier position is important in the case of {\it Umlautung}, 
such as the change of e.g. {\it a} $\rightarrow$ {\it \"a} between different inflected forms, 
which is a non-concatenative process that cannot 
be modeled by segmentation-based approaches such as BPE or our knowledge-poor approach.
This idea will be further discussed in Section~\ref{sec:discussion}, based on the examples given in
Tables~\ref{table:bpe-splittings-schweigen} and \ref{table:bpe-splittings-straftatbestand}.


\section{Knowledge-poor approach: The word segmentation strategy}
\label{sec:target-side-segmentation}

The knowledge-poor approach splits words into smaller sub-word units. 
Previous approaches to sub-word NMT have followed the same core principle, 
which limits the set of symbols known to the model, while at the same time
allowing for open-vocabulary machine translation. 
A prominent example is BPE word segmentation. 

Whereas BPE is purely frequency-driven, we strive to integrate a basic amount
of linguistic supervision in order to model target-side morphology. BPE
sub-words are often not linguistically sound, but we want the machine
translation system to reliably inflect output words.  The neural model should
be able to learn morphological word formation processes from its training data.
Our goal is therefore to provide the model with better morphological guidance
through a more linguistically informed word segmention on the target side of
the training corpus.

The first idea behind linguistically
informed word segmentation is to separate inflectional suffixes from word
stems. The segmentation strategy relies on the very same manually defined
suffix detection rules that are in wide-spread use for other applications in
Natural Language Processing and Information Retrieval that benefit from word stemming. 
Secondly, the approach targets the issue of productive compounding, a
common process in many Germanic languages, by integrating a compound splitter. 
Suffix splitting and compound splitting are combined in the overall pipeline
for linguistically informed word segmentation and cascaded with BPE. 

\begin{table}[t!]
\caption{German suffixes that the suffix splitter can separate from a word stem.}
\centering
\small
\begin{tabular}{c}
\hline \hline
{\bf Suffixes}\\
\hline
\tt{\mbox{-e}}, \tt{\mbox{-em}}, \tt{\mbox{-en}}, \tt{\mbox{-end}}, \tt{\mbox{-enheit}}, \tt{\mbox{-enlich}}, \tt{\mbox{-er}}, \tt{\mbox{-erheit}}, \tt{\mbox{-erlich}}, \tt{\mbox{-ern}}, \tt{\mbox{-es}}, \\
\tt{\mbox{-est}}, \tt{\mbox{-heit}}, \tt{\mbox{-ig}}, \tt{\mbox{-igend}}, \tt{\mbox{-igkeit}}, \tt{\mbox{-igung}}, \tt{\mbox{-ik}}, \tt{\mbox{-isch}}, \tt{\mbox{-keit}}, \tt{\mbox{-lich}}, \\
\tt{\mbox{-lichkeit}}, \tt{\mbox{-s}}, \tt{\mbox{-se}}, \tt{\mbox{-sen}}, \tt{\mbox{-ses}}, \tt{\mbox{-st}}, \tt{\mbox{-ung}} \\
\hline
\end{tabular}
\label{tab:affixes}
\end{table}

\subsection{Linguistically informed word segmentation pipeline}
\label{sec:segmentation-pipeline}

In detail, the overall cascaded pipeline for linguistically informed word
segmentation consists of three steps:

\begin{enumerate}
\item A suffix splitter is applied that separates common German
  morphological suffixes from the word stems. 
  The suffix splitter is a modification of the German Snowball stemming
  algorithm from NLTK.\footnote{\url{http://www.nltk.org/_modules/nltk/stem/snowball.html}} 
  As opposed to usual stemming scenarios \citep{Porter1980}, suffixes are
  however not discarded, but kept as a detached token; also, different
  from the stemming algorithm, no modifications are applied to the stem
  part of the word, such as lowercasing or Umlaut replacement ({\it \"a}, 
  {\it \"o}, {\it \"u} to {\it a}, {\it o}, {\it u}, etc.). 
  The German Snowball stemming algorithm may internally identify multiple
  consecutive suffixes of a word, which we decide to keep as separate tokens.
  For instance, {\it wirtschaftlichen} ({\it economical}) is segmented into {\tt wirtschaft
  \$\$lich \$\$en} rather than {\tt wirtschaft \$\$lichen}. (The {\tt \$\$}
  characters are a special marker that we add.)
  Table~\ref{tab:affixes} lists German suffixes that the suffix splitter can
  detach from a word stem.
\item Next, the empirical compound splitter as described by \citet{E03-1076} is applied.
  A Perl implementation is part of the Moses toolkit \citep{koehn-EtAl:2007:PosterDemo}. 
  We choose an aggressive configuration of the compound
  splitter (\texttt{-min-size 4 -min-count 2 -max-count 999999999}) 
  in order to end up with a relatively small token vocabulary. 
  We prevent the compound splitter from segmenting suffix tokens that were
  separated in the previous step. 
  We also introduce a minor modification as compared to the Moses compound
  splitting script in standard settings. 
  The standard settings take the filler letters ``{\it s}'' and
  ``{\it es}'' into account, which often appear between word parts in
  German noun compounding. 
  For better consistency of the compound splitting component with suffix
  splitting, we additionally allow for more fillers, namely: suffixes,
  suffixes followed by ``{\it s}'', and ``{\it zu}''.
\item The BPE technique is finally applied on top of the suffix-split and
  compound-split data in order to further reduce the vocabulary size.  This
  last step is conducted only for efficiency reasons in NMT. 
  Suffix splitting and compound splitting alone are not suitable for arbitrary
  reduction of the vocabulary size. 
  We use ``joint'' BPE in this work, i.e., the BPE merge operations are learned on a
  concatenation of the target and source language side of the parallel
  training corpus. 
\end{enumerate}

\begin{table}
\caption{Declension of the German noun ``Fisch'' (English: ``fish'') in
  singular and plural in all four German cases. For this example, the singular
  genitive and dative cases allow for two valid alternatives each.}
{\small
\setlength\tabcolsep{1.0em}
\begin{tabular}{lll}
\hline
\hline
            & {\bf Singular}  & {\bf Plural} \\
\hline
  {\bf Nominative}  & {\tt Fisch}           & {\tt Fisch\underline{e}} \\
  {\bf Genitive}    & {\tt Fisch\underline{es}/Fisch\underline{s}}  & {\tt Fisch\underline{e}} \\
  {\bf Dative}      & {\tt Fisch/Fisch\underline{e}}    & {\tt Fisch\underline{en}} \\
  {\bf Accusative}  & {\tt Fisch}           & {\tt Fisch\underline{e}} \\
\hline
\end{tabular}
}
\label{table:cases}
\end{table}

\subsection{Example}

To give an example of how the word segmenter operates, we present in
Table~\ref{table:cases} all declensions of a German noun, {\it Fisch} ({\it fish}). 
The suffix splitting component of the word segmenter separates the
underlined suffixes from any appearence of an inflected occurence of that noun
in the training corpus. The dative plural German noun {\it Fischen} becomes
{\tt Fisch \$\$en}.  We insert a space character between the stem and the
suffix, and the suffix is prepended with an attached special indicator {\tt
\$\$} to enable reversibility, an important aspect for post-processing of the
NMT system's output.  Assuming that inflected forms of {\it Fisch} are present
a couple of times in the training corpus, we can hope for the neural model to
learn the word's inflectional variants from the training data. We also
counteract data sparsity, because the stem {\tt Fisch} is now a separate token
that likewise appears in all word-segmented versions of training instances that
contained any of the declensions of the noun. Furthermore, if some inflected
variant of a different noun was unobserved, but that noun follows a regular
inflection pattern, the neural model can in principal be able to generate an
unseen combination of stem and suffix. For example, if the dative plural variant of
the noun {\it Tag} was unobserved, but the model knows from sentence context
that an English input word {\it day} needs to be translated to a dative plural
form, then it can produce the output sequence of tokens {\tt Tag \$\$en},
resulting in a correct morphological form. Finally, we think that the detached
suffix tokens facilitate learning
how to produce correct linguistic
agreement between output words.

The second cascaded component of the linguistically informed word segmentation
pipeline is the compound splitter.
We explain the utility of compound splitting
by following up on
the {\it Fisch} example. There exist different types of fish, such as {\it
ornamental fish} or {\it juvenile fish}. These are often expressed as compound
in German, e.g.\ {\it Zierfisch} or {\it Jungfisch}. Each of
these compounds
can
be inflected, e.g.\ {\it Zierfischen} -- {\tt Zierfisch \$\$en} after suffix
splitting. Compound splitting gives us {\tt \#U zier @@ Fisch \$\$en}, where
``{\tt @@}'' is a standalone compound separator token that we introduce, and
``{\tt \#U}'' indicates that uppercasing is required when the compound is
re-merged from the sub-words (``{\tt \#L}'' for lowercasing).  The model can
now learn to produce new compounds at inference time, such as {\tt \#U zier @@
Gegenstand} ({\it ornamental object}).

\subsection{Limitations}

For the input sentence {\it the EU commission wants to double the limits for
mercury in large predatory fish} from
Table~\ref{table:lemma-tag-example}, a neural model that was trained with
linguistically informed target-side word segmentation outputs the translation: 
{\tt die EU @-@ Kommission möcht \$\$e die \#U Grenz @@ Wert \$\$e für Quecksilb \$\$er in groß \$\$en \#U Raub @@ Fisch \$\$en verdoppeln}.

While this example translation is fluent and adequate, it also highlights
drawbacks of the knowledge-poor approach. 

The conjugated verb {\it möchte} is properly split into stem and suffix, but
the verb is irregular. The knowledge-rich approach with full morphological
analysis would know the lemma {\it mögen} and combine it with a morphological
feature tag, which the morphological generation tool would employ to map the
lemma-tag pair to the surface form {\it möchte} in post-processing. The
knowledge-poor segmentation strategy generalizes less well over inflectional
variants. This holds not only for verbs, but also for nouns, where the stem may
occasionally be altered with a change in case or number, such as {\it Haus}
({\it house}) being {\it H\"auser} in plural, or the similar {\it
Umlautung} in {\it B\"aume$_{Pl}$ $\leftrightarrow$ Baum$_{Sg}$} that we 
mentioned in Section~\ref{sec:generation-lemma-tag} already.

The noun {\it Quecksilber} ({\it mercury}/{\it quicksilver}) is
segmented because the stemming algorithm's simple rules have failed to
recognize that the suffix is not inflectional in this instance. The system has
produced the right output word from its parts, but such flawed splits in the
training data most likely hamper the learning of inflectional patterns.

Some shortcomings of our current pipeline would vanish with improved stemming
algorithms \citep{weissweiler17:gscl} and compound splitting tools. We however
believe that the knowledge-rich lemma-tag strategy has some conceptual
advantages over plain word segmentation. 
The linguistically informed word segmentation strategy, in turn, will typically
be implementable much more quickly for new languages than the knowledge-rich
lemma-tag strategy, because coding stemmer-like rules is much simpler than
building a full-fledged morphological analyzer, lemmatizer, and generation tool.
Depending on the runtime efficiency of the morphological analysis tool, the
training data preparation can also take longer for the lemma-tag strategy.


\section{Empirical evaluation: Machine translation experiments}
\label{sec:experiments}
In this section, we present and discuss the results of the two strategies, first in 
a general-language setting, and then in a cross-domain experiment translating medical data.
Additionally, both strategies are repeated with another translation toolkit, 
the {\it Nematus} system.



\subsection{Experimental setup}

\subsubsection{Data}
An important question for the evaluation of the two presented strategies is the performance on
data sets of different sizes and domains. For the performance on general language, we look at
three settings: a small corpus (248,730 parallel sentences), a medium-sized corpus (1M parallel 
sentences) and a large corpus (1,956,444 parallel sentences), where the small corpus consists of 
the news-commentary data set, the large corpus combines Europarl with the news-commentary corpus, 
and the medium corpus is a random subset of Europarl combined with the news-commentary corpus 
(after filtering, see below). 
As development and test sets, we use the WMT'15 (dev/validation) and WMT'16 (test) news\-test sets.%
\footnote{All corpora are freely available from the WMT shared tasks website at
\url{http://www.statmt.org/wmt19/}. Shared task \BLEU scores can be viewed on
the WMT evaluation matrix website at \url{http://matrix.statmt.org/}.
\citet{bojar-etal-2015-findings} and \citet{bojar-etal-2016-findings} provide
the official WMT'15 and WMT'16 shared task results including human evaluation.}

The presented approaches affect the sentence length: the lemma-tag approach essentially doubles the
sentence length by inserting tags, in addition to BPE splitting applied to the tag-lemma pairs; the
target-side word segmentation approach also leads to considerably longer sentences than the standard
BPE splitting.
For efficiency reasons during training, it is common practice to set a maximum sequence length; training sentences
longer than that are lost (or partially lost) during training. For this reason, the increased sentence length
in the lemma-tag and the segmentation strategy needs to be addressed in the setup of the training data.
To ensure that both the baseline system and the linguistically informed systems can make use of all 
training sentences, the data is prepared in a way that the maximum sentence length can just be set to a high 
value (or the longest sentence occurring in the corpus) to include all sentences in the training.
This is achieved by restricting the sentence length, including BPE segmentation which can result in considerably
increased sentence lengths in some cases.
To avoid overly long sentences, the training data was first filtered to sentences of length 50, and in 
a second filtering step, all sentences containing more than 60 words after standard BPE
splitting were removed. This second filtering targeted sentences containing mostly foreign
language words or characters, being split nearly at character level. 
Length filtering was only applied to the training data, but not to the development and test
data.

\paragraph{Data pre-processing: Baseline.}
The baseline is a system trained on standard surface forms (tokenized and truecased),
segmented with BPE to the configured vocabulary size.
The English and German sides were concatenated prior to BPE segmentation in order to have a consistent
representation for shared vocabulary between source and target side (also known as ``joint'' BPE).

\paragraph{Data pre-processing: Lemma-tag strategy.} 
The training data for the lemma-tag strategy was prepared based on data parsed with BitPar 
\citep{schmid04:coling}
for the morphological features to be annotated to the tags.  The lemmas were obtained through analysis with 
SMOR \citep{schmid04:lrec}, as described in Section~\ref{sec:training-data-lemma-tag}.
 
While reducing inflected word forms to lemmas and tags decreases the vocabulary of the German data,
a further reduction by means of BPE is still required: the lemma-tag strategy addresses only inflected forms,
i.e. leaving other words, such as proper names, unmodified. Furthermore, the English side is not affected by
the lemma-tag modification.
Prior to training the translation model, the German and English side are thus segmented with BPE until 
the desired vocabulary size is reached (29,500 merge operations). 
As for the baseline experiments, the English and German sides are concatenated to enable a consistent representation 
of shared vocabulary. The abstract representation in SMOR format should not pose a problem, as shared vocabulary 
mostly consists of named entities that are not subject to the lemmatized representation anyway, but remain in their 
original format. 
 
\paragraph{Data pre-processing: Segmentation strategy.} 
For the setup with linguistically informed target word segmentation,
tokenization and truecasing were the same as for the baseline, except that we
additionally applied hyphen splitting on both the source and the target side,
which for instance turns {\tt  EU-Kommission} into {\tt  EU @-@ Kommission}. 
We then applied the full word segmentation pipeline as outlined in detail in
Section~\ref{sec:segmentation-pipeline} to the German target-language side of
the training data. The English source side was BPE-split with the ``joint'' BPE
model. The number of BPE merge operations was set to the same amount that had
also been configured for the baseline and for the lemma-tag setup (29,500 merge
operations). Also consistent with all other setups in this study, we attach the
marker that indicates a BPE segmentation point (for this setup: {\tt \#\#})
to the right end of the sub-word that appears to the left of the segmentation
point.  Compound split indicators remain standalone symbols (as in {\tt  \#U
Stahl @@ Werk}; or, with a filler, {\tt \#U Jahr @es@ Wechsel}), which we found to
be important for translation quality. The three cascaded splitters can segment
single words into fairly long sub-word sequences, such as the sequence {\tt \#U
Neben\#\# erwerb @s@ Land @@ Wir\#\# t \$\$e}.

\subsubsection{Transformer system configuration}
The experiments were carried out using a {\it Transformer} NMT model with the Sockeye toolkit \citep{Sockeye:17}. 
Table~\ref{table:transformer-parameters} shows the training hyperparameters. 
Additional Sockeye configuration options that are not listed have been kept at their defaults. 
Our configuration settings are a mix of conventional values
\citep{vaswani2017attention,Sockeye:17} that we have had good experience with
when previously building competitive machine translation systems
\citep{huck-EtAl:2018:WMT}.

\begin{table}
\caption{Sockeye hyperparameter settings for the Transformer model.}
{\small
\setlength\tabcolsep{0.5em}
\begin{minipage}{\textwidth}
\begin{tabular} {lllll}
\hline
\hline
  encoder & transformer          &~~~~&   num-layers & 6 \\
  decoder & transformer          &&       label-smoothing & 0.1 \\
  batch-type & word              &&       transformer-dropout-act & 0.1 \\
  batch-size & 4096              &&       transformer-dropout-attention & 0.1 \\
  initial-learning-rate & 0.0002 &&       transformer-dropout-prepost & 0.1 \\
  max-seq-len & 200              &&       checkpoint-frequency & 3000 \\
\hline
\end{tabular}
\end{minipage}
  }
\label{table:transformer-parameters}
\end{table}

\subsection{Experimental results}

\subsubsection{In-domain translation with a Transformer system}

Table~\ref{table:results-transformer} shows the results for the baseline, i.e. a system with
standard BPE splitting, the lemma-tag system and the system applying target-side segmentation.
While the lemma-tag and the segmentation strategies outperform the baseline when trained on
the small data set, their performance is at the same level as the baseline for the medium and 
the large setting.\footnote{Significance was computed with the script {\tt bootstrap-hypothesis-difference
-significance.pl} that is part of the Moses package, available from \url{https://github.com/moses-smt/mosesdecoder/}.}

\begin{table}
\caption{Results with Transformer for the LemmaTag and Segmentation approaches in comparison to a system trained on surface forms (Baseline) in case-sensitive \BLEU.
Significant improvements over the baseline (p-value=0.05) are marked with~*.}
{\small 
\begin{minipage}{\textwidth}
\setlength\tabcolsep{0.5em}
\begin{tabular} {llll}
  \hline
  \hline
  \multicolumn{4}{c}{\bf English--German In-domain Translation} \\
  \hline
           & {\bf Small}  & {\bf Medium} & {\bf Large} \\
  \hline  
    {\bf Transformer BPE Baseline} & \rdm{21.48}  & \rdm{27.40} & \rdm{28.96} \\     
    {\bf Transformer LemmaTag}     & \rdm{21.89}\,*  & \rdm{27.49} & \rdm{28.83}  \\       
    {\bf Transformer Segmentation} & \rdm{22.18}\,*  & \rdm{27.00} & \rdm{29.00}  \\
  \hline 
\end{tabular}
\end{minipage}
  }
\label{table:results-transformer}
\end{table}


\subsubsection{Out-of-domain translation with a Transformer system}

Many translation scenarios involve the handling of low-resource data, where the main difficulty lies
in setting up a translation model with only little available domain-specific training data, if at all.
In such a situation, the problems caused by rich (target-side) morphology are typically aggravated,
as inflectional variants are less likely to appear in the limited amount of training data and thus cannot
be learned and produced by the system.

A domain that differs greatly from general language is the medical domain which has an obvious difference 
in the used vocabulary.
Applying a system trained on general language, but with a component to handle target-side morphology thus 
constitutes an interesting use case. For this experiment, we use a test set
from the project {\it HimL} (Health in my Language) \citep{haddow17:eamt:himl}
consisting of 1931 sentences.%
\footnote{\url{http://www.himl.eu/files/himl-test-2015.tgz}}
This test set consists of data extracted from {\it NHS 24} (the {\it National Health Service}) and 
{\it Cochrane}\footnote{\url{https://www.cochrane.org/}} online content.
While the {\it NHS} data contains health information aimed at the general public, the {\it Cochrane} part 
consists of summaries of scientific studies, and differs considerably from the NHS-based sentences.
The German reference translations were obtained
by post-editing, with the initial automatic translation 
created by a Moses phrase-based MT system. 
Note that this test set has also been standardly used in the biomedical shared
tasks at WMT \citep{jimeno-yepes-etal-2017-findings,neves-etal-2018-findings},
but there large biomedical training data sets were made available, in contrast
to our study, which instead looks at the difficult out-of-domain translation
task.

\begin{table}
  \caption{Results for the out-of-domain medical test set for the LemmaTag and
  Segmentation approaches in comparison to a system trained on surface forms
  (Baseline) in case-sensitive \BLEU. Significant improvements over the baseline (p-value=0.05) are marked with~*.} 
{\small
\begin{minipage}{\textwidth}
\setlength\tabcolsep{0.5em}
\begin{tabular} {llll}
  \hline
  \hline
  \multicolumn{4}{c}{\bf English--German Out-of-domain Translation} \\
  \hline
           & {\bf Small} &  {\bf Medium} & {\bf Large} \\
  \hline  
    {\bf Transformer BPE Baseline} & \rdm{18.00} & \rdm{23.31} & \rdm{24.40} \\     
    {\bf Transformer LemmaTag}     & \rdm{19.01}\,* & \rdm{24.20}\,* & \rdm{25.09}\,*  \\       
    {\bf Transformer Segmentation} & \rdm{18.93}\,* & \rdm{22.89} & \rdm{24.75}  \\
  \hline
\end{tabular}
\end{minipage}
  }
\label{table:results-transformer-himl}
\end{table}


Table~\ref{table:results-transformer-himl} gives the results for translating data from the medical
domain.
The lemma-tag strategy as well as the segmentation approach lead to improved
results over the baseline system, particularly with the Small training size.
Interestingly, the lemma-tag strategy also provides for statistically significant gains for the Medium and Large training sizes.
The gains are larger because
the translation task is more difficult and thus can benefit more
easily from the linguistic information, and the ability to generate new words and word forms.
Table \ref{table:example-himl} in section \ref{subsection:out-of-domain} discusses an example 
translation from the medical domain.

\subsection{Sanity check: Comparison with a shallow-RNN system}
Our previous work showed larger gains when working with shallow-RNN systems for the lemma-tag approach.
For this reason, we decided to run experiments applying both strategies using a shallow-RNN translation model 
and examine these as well.

\subsubsection{Shallow-RNN system configuration}
The RNN experiments were carried out using the Nematus toolkit \citep{nematus}.
The RNN is shallow, i.e., we use one single hidden layer, not a deep model. 
Like in the Transformer experiments with Sockeye, for Nematus we again settle on a
minor variation of configuration settings that we had already employed in
previous work \citep{huck-braune-fraser:2017:WMT} with a top shared task result
\citep{bojar-EtAl:2018:WMT1}.
Table~\ref{table:nematus-parameters} lists the Nematus hyperparameters used in
this set of experiments. Additional Nematus configuration options that are not
listed were kept at their default values. 
We trained with the Adam optimizer \citep{KingmaB14}, a batch
size of 128, and, due to memory issues with the increased sentence length, 116
for the lemma-tag experiments.

\subsubsection{In-domain translation with a shallow-RNN system}

Table~\ref{table:results-nematus} shows the results of in-domain translation with shallow-RNN systems.
In contrast to the experiments with 
Transformer, the lemma-tag approach is able to outperform the baseline system for both the Small and Large training data sizes, even though the \BLEU difference 
decreases with increasing training data size. This outcome is not surprising, as linguistic
information tends to become less effective if more training data is available.
\footnote{For example, \cite{Burlot16reinflection} present
 a set of experiments with varying training data sizes for a task similar to ours, and report that 
 they see less impact of linguistic  modeling with increased training data.
 The general reasoning behind this observation is that more training data means that the system sees 
 more word forms during training, and thus is able to derive better statistics. As a result, the 
 linguistic modeling, i.e.\ better generalization, becomes less effective.}
However, even in the setting with nearly 2M parallel sentences, the lemma-tag system still benefits.
The results for the segmentation system are again more mixed.

\begin{table}
\caption{Nematus hyperparameter settings for the shallow-RNN model.}
{\small
\begin{minipage}{\textwidth}
\setlength\tabcolsep{0.5em}
\begin{tabular}{lllll}
\hline
\hline
                     &&~~~~& dropout & yes \\
vocab size & 30k          && dropout embedding & 0.2 \\
embedding size & 500      && dropout hidden & 0.2 \\
hidden layer size & 1024  && dropout source & 0.1 \\
learning rate & 0.0001    && dropout target & 0.1 \\
\hline
\end{tabular}
\end{minipage}
}
\label{table:nematus-parameters}
\end{table}

\begin{table}
\caption{Results with {\it Nematus} for the LemmaTag approach in comparison to a system trained on surface forms (Baseline) in case-sensitive \BLEU. Significant improvements over the baseline (p-value=0.05) are marked with~*.}
{\small
\begin{minipage}{\textwidth}
\setlength\tabcolsep{0.5em}
\begin{tabular} {llll}
  \hline
  \hline
  \multicolumn{4}{c}{\bf English--German In-domain Translation} \\
  \hline
           & {\bf Small} & {\bf Medium} & {\bf Large} \\
  \hline  
    {\bf Shallow-RNN BPE Baseline} & \rdm{22.14} & \rdm{26.39} & \rdm{27.54} \\     
    {\bf Shallow-RNN LemmaTag}     & \rdm{23.35}\,* & \rdm{26.73} & \rdm{28.26}\,* \\       
    {\bf Shallow-RNN Segmentation} & \rdm{22.42} & \rdm{26.65} & \rdm{27.75} \\
  \hline
\end{tabular}
\end{minipage}
  }
\label{table:results-nematus}
\end{table}


\subsubsection{Out-of-domain translation with a shallow-RNN system}

Table~\ref{table:results-nematus-himl} shows the results for the surface system and the lemma-tag system: 
As in the Transformer experiments, the lemma-tag system achieves better results in \BLEU for all settings. 

\begin{table}
\caption{Results for the LemmaTag approach in comparison to a system trained on surface forms (Baseline) in case-sensitive \BLEU for medical data (HimL test set). Significant improvements over the baseline (p-value=0.05) are marked with~*.}
{\small
\begin{minipage}{\textwidth}
\setlength\tabcolsep{0.5em}
\begin{tabular} {llll}
  \hline
  \hline
  \multicolumn{4}{c}{\bf English--German Out-of-domain Translation} \\
  \hline
           & {\bf Small} & {\bf Medium} & {\bf Large} \\
  \hline  
    {\bf Shallow-RNN BPE Baseline} & \rdm{19.33} & \rdm{24.10} & \rdm{25.39} \\    
    {\bf Shallow-RNN LemmaTag}     & \rdm{19.98}\,* & \rdm{24.23} & \rdm{25.98}\,* \\ 
    {\bf Shallow-RNN Segmentation} & \rdm{19.20} & \rdm{23.92} & \rdm{24.90} \\    
  \hline
\end{tabular}
\end{minipage}
  }
\label{table:results-nematus-himl}
\end{table}


\section{Discussion of the linguistic impact and examples:\\ BPE splitting vs. morphologically informed modeling}
\label{sec:discussion}
The experiments in the previous section showed that both the lemma-tag strategy and the segmentation strategy
can improve the translation quality, even though the impact decreases with larger training data, in particular when
looking at translating news data as opposed to special domains such as medical data.
In this section, we try to get some insight into the effects on the linguistic level, in particular the
effects of BPE splitting in contrast to the linguistically informed approaches, and compare the data representation 
of the different strategies.

BPE \citep{sennrich-bpe:acl2016} is a common technique to reduce the vocabulary size in NMT. It 
relies entirely on word and sub-word frequencies observed in the data to split and thus does not require any 
external resources. However, the fact that it is not linguistically guided leads to sub-optimal splitting.

\begin{table}
\caption{Representation of inflection variants of the verb ``schweigen'' (to remain silent) 
in the training data of the large baseline system. Inflectional suffixes are highlighted in the first column.}
{\small
\begin{minipage}{\textwidth}
\begin{tabular} {llll}
  \hline
  \hline {\bf Word} & {\bf Freq} & {\bf BPE} & {\bf Comment}\\
  \hline
     schweig{\bf en}   &  763 & schweigen      & infinitive, present 3rd person plural\\  
     schweig{\bf t}    &   78 & schwei@@ gt    & present 3rd person singular\\
     schweig{\bf e}    &    1 & schwei@@ ge    & conjunctive 3rd person singular\\
     schwieg{\bf en}   &    9 & sch@@ wiegen   & past 3rd person plural\\
     schwieg{\boldmath $\emptyset$}     &    8 & sch@@ wie@@ g  & past 3rd person singular\\
     {\bf ge}schwieg{\bf en} &   55 & gesch@@ wiegen & past participle\\
  \hline 
\end{tabular}
\end{minipage}
  }
\label{table:bpe-splittings-schweigen}
\end{table}

\begin{table}
\caption{Representation of inflection variants of the noun ``Straftatbestand'' (criminal offence) 
in the training data of the large baseline system. Inflectional suffixes are highlighted in the first column.}
{\small
\begin{minipage}{\textwidth}
\begin{tabular} {llll}
  \hline
  \hline {\bf Word} & {\bf Freq} & {\bf BPE} & {\bf Comment}\\
  \hline
   Straftatbeständ{\bf en} & 16 &   Straft@@ at@@ best@@ änden & plural, dat \\
   Straftatbeständ{\bf e}  & 40 &   Straft@@ at@@ bestände & plural, acc/nom/gen\\
   Straftatbestand{\boldmath $\emptyset$}  & 64 &   Straft@@ at@@ bestand & singular, acc/nom/dat\\
   Straftatbestand{\bf s}  &  6 &  Straft@@ at@@ best@@ ands & singular, gen\\
  \hline 
\end{tabular}
\end{minipage}
  }
\label{table:bpe-splittings-straftatbestand}
\end{table}


The example in Table~\ref{table:bpe-splittings-schweigen} lists inflectional variants of the verb
{\it schweigen} ({\it to remain silent}) which illustrate several problems that arise when using BPE on surface data:
\begin{itemize}
  \item Non-concatenative processes, such as the shift from {\it ei} in present tense to {\it ie} in
        past tense, cannot be captured. Such changes of vowels in the stem are common in many German verbs and nouns.
  \item Inconsistent splitting of inflectional suffixes (here, there is no splitting in the infinitive form, 
        but an approximative splitting of inflectional suffixes in two of the present tense forms ({\it schweigt, 
        schweige})). 
  \item Different splitting for past tense forms, where the suffix {\it -en} remains attached, but the stem 
        is split in the middle. 
        The resulting segment {\it sch@@} is rather meaningless and, being a frequent German n-gram, can be
        found in many different contexts. Even worse, the part {\it wiegen} is another, unrelated word 
        ({\it wiegen} = {\it to weigh})
        which can
        lead to confusion with actual occurrences of {\it wiegen} 
        during training.
\end{itemize}

\noindent
It becomes clear from looking at the different forms that they are not represented efficiently by BPE. 
In the lemma-tag approach, the lemma is simply represented as {\tt schweigen<V>} in all instances, 
accompanied by the respective morphological tag.

Similar problems are shown in Table~\ref{table:bpe-splittings-straftatbestand} for inflectional variants of the 
complex noun {\it Straftatbestand} ({\it criminal offence}). Again, the inflectional suffixes are not handled in a 
consistent way. Furthermore, the first part of the compound is split unintuitively into {\it Straft+at},
rather than into {\it Straf+tat} ({\it punishable+deed}).
In the lemma-tag system, the lemma is represented as {\tt strafen<V>@@ Tatbestand}\footnote{The splitting into 
{\it strafen+Tatbestand} vs.\ {\it Straftat+Bestand} is questionable, but it is consistent over all forms.}
after BPE splitting.

The reduction of surface forms by replacing inflected forms with lemma-tag pairs leads to a considerable decrease
in surface forms, which provides a better basis for BPE splitting. Similarly, the segmentation strategy results in
a linguistically sound and overall consistent segmentation, such that the subsequent BPE step will find compounds
and inflectional suffixes already split. 

Table~\ref{table:vocabulary-size} contrasts the vocabulary sizes in the respective settings and system variants.
While the lemma-tag strategy already reduces the vocabulary size, the segmentation strategy leads to a considerable
further reduction, presumably due to the explicit compound splitting prior to BPE.

\begin{table}
\caption{Comparison of German vocabulary size for the different corpus settings.}
{\small
\begin{minipage}{\textwidth}
\begin{tabular} {lccc}
  \hline
  \hline {\bf Corpus} & {\bf Baseline} & {\bf LemmaTag} & {\bf Segmentation}\\
  \hline
  Small   & 159,908  & 117,386 & \phantom{0}57,715 \\
  Medium  & 300,224  & 227,292 & \phantom{0}92,433 \\
  Large   & 401,256  & 306,190 & 113,966 \\  
  \hline 
\end{tabular}
\end{minipage}
  }
\label{table:vocabulary-size}
\end{table}

\section{Translation examples}
\label{section:translation-examples}
In this section, some example translations are shown and discussed, in particular with regard to the
creation of new words. The last example serves as basis to discuss the impact of the linguistic
approaches on the morphological level.

\subsection{Creation of new words}
The creation of new words based on linguistically sound parts 
(segmentation strategy) or by means of
word generation (lemma-tag strategy) is an important factor in these translation approaches.
In the following, we discuss two examples containing words not seen in the training data.

Table~\ref{table:example-kirchturm} shows an example for the creation of a new word as a translation for 
{\it church tower}: {\it Kirchturms} in the lemma-tag system, and {\it Kirchenturms} in the baseline output. 
The variant {\it Kirchturms} in the lemma-tag system is correct (cf.\ reference translation); 
the variant in the baseline is understandable, but the realization of the transitional element is incorrect. 
The segmentation system is making the same mistake. 
While there are instances of {\it Kirchturm} and {\it Kirchtürme} (plural) in the training data, the genitive
form {\it Kirchturms} is unseen. For the lemma-tag version, it is rather straightforward to generate the 
respective form.

\begin{table}
\caption{Example: outputs of the baseline Transformer system (B) in comparison to
  the lemma-tag (LT) and segmentation (S) Transformer systems.}
{\small
\begin{tabular} {ll}
\hline
\hline
{\bf B}   & Neben den Resten der Festung und des {\bf Kirchenturms} \ldots \\
\hline
{\bf LT}  & Neben den Überresten der Festung und des {\bf Kirchturms} \ldots \\
\hline
{\bf S}   & Neben den Überresten der Festung und des {\bf Kirchenturms} \ldots \\
\hline
{\bf REF} & Neben den Überresten der Festung und des {\bf Kirchturms} \ldots \\
\hline%
{\bf SRC} & Besides the remains of the fortress and the {\bf church tower} \ldots \\
\hline
\end{tabular}
}
\label{table:example-kirchturm}
\end{table}


The example in Table~\ref{table:example-himl-transformer} contrasts a sentence pair from the medical domain 
where the lemma-tag output contains the newly created word 
{\it patientenrelevanten} ({\it patient-relevant}), generated from the
tag-lemma pair {\tt <+ADJ><NoGend><Gen><Pl><Wk> Patient<NN>@@ relevant<Pos>}, with {\tt @@} marking a BPE
segmentation point.
In contrast, the baseline produced the non-existing form {\it patientenwichtigen}, obtained from the 
BPE segments {\it pati@@ ent@@ en@@ wichtigen} -- while this word creation can be easily understood, it is
not the correct term in this context.
We cannot definitely say why the the lemma-tag system generated the correct term, 
given that there is no instance of {\it patientenrelevant} at all in the training data.
However, one possibility might be that the lemma-tag system learned that the structure {\it Word + ``relevant''} more
frequently leads to a valid adjective than {\it wichtig} ({\it important}) - hence, the more structured
representation could have enabled indirectly the generation of the correct form.
The different aspects and possibilities of word formation in the lemma-tag system are very interesting,
and we plan on studying this task in future work.

The system with linguistically informed target word segmentation receives the
hyphen-split input word {\it patient @-@ important} from its pre-processing and,
in this context, translates it to the output token sequence {\it \#L Patient @en@
relevant \$\$en}, which is assembled to the valid German word {\it
patientenrelevanten} in post-processing.

When comparing the translation outputs with the reference translation, it becomes clear that the
improvement in the lemma-tag system and the segmentation system is not reflected in the \BLEU score, as
the sentence is phrased differently in the reference translation. This is a common problem with \BLEU 
(and most other automatic metrics) as they rely on matches in a reference translation. 

\begin{table}
    \caption{Example for creating a new word: outputs of the baseline system (B) in comparison to the lemma-tag system (LT) and the segmentation system (S).}
{
\centering
{\small
\begin{tabular} {ll}
\hline
\hline
{\bf B}   & um Verbesserungen an {\bf patientenwichtigen} klinischen Ergebnissen zu erkennen \\
          & {\it to recognize improvements in ``patient-important'' clinical outcomes} \\
\hline
{\bf LT}  & um Verbesserungen im Bereich der {\bf patientenrelevanten} klinischen Ergebnisse \\
          & zu erkennen \\
          & {\it to recognize improvements in the area of patient-relevant clinical outcomes} \\
\hline
{\bf S}   &  um Verbesserungen bei {\bf patientenrelevanten} klinischen Ergebnissen \\
          &  festzustellen \\
          & {\it to determine improvements in patient-relevant clinical outcomes} \\
\hline
{\bf REF} & um bessere {\bf f\"ur die Patienten wichtige} klinische Ergebnisse zu erreichen \\
          & {\it to obtain better outcomes that are important to the patients} \\
\hline
{\bf SRC} & to detect improvements in {\bf patient-important} clinical outcomes \\
\hline
\end{tabular}
}
}
\label{table:example-himl-transformer}
\end{table}


\subsection{Looking at the morphological level}

\begin{table}
\caption{Example: outputs of the baseline Nematus system (B) in comparison to the lemma-tag (LT) 
and segmentation (S) Nematus systems.}
{
\centering
{\small
\begin{tabular} {p{0.6cm}l}
\hline
\hline
{\bf B}   & {\bf Dieser Wertpapierkonto} wurde im Namen der beiden Angeklagten gef\"uhrt , \\
          & und laut einer Erkl\"arung des Clubs war seine Existenz dem Club nicht bekannt . \\
\hline
{\bf LT}  & {\bf Dieses Wertpapierkonto} wurde im Namen der beiden Angeklagten gef\"uhrt , \\
          & und laut einer Erkl\"arung des Clubs war seine Existenz dem Club nicht bekannt . \\
\hline
{\bf S}   & {\bf Diese Wertpapierkonten} wurden im Namen der beiden Angeklagten gef\"uhrt , \\
          & und laut einer Erkl\"arung des Clubs war ihre Existenz dem Club unbekannt . \\
\hline
{\bf REF} & {\bf Dieses Wertpapierkonto} lief auf den Namen des zweiten Angeklagten \\
          & und war laut dessen Aussage dem Verein nicht bekannt . \\
\hline
{\bf SRC} & {\bf This securities account} was run in the name of the two defendants and \\
          & according to a statement by the club its existence was not known to the club . \\
\hline
\end{tabular}
}
}
\label{table:example-lem-tag}
\end{table}

When comparing the outputs of the different systems, it is difficult to find systematic differences.
Obvious errors on the morphological level such as wrong agreement are rare, but they do exist, as can
be seen in the example in Table~\ref{table:example-lem-tag}:  
the translations are identical and correct, except for the inflection of {\it dieser} ({\it this}) in 
the baseline output. Even though the following word, {\it Wertpapierkonto} ({\it securities account}, 
literally {\it securities paper account}) does not exist in the training data, it has been produced 
by both systems. 
In the lemma-tag system, the generation step has been straightforward: the word is composed through a BPE
operation (marked by {\tt @@}) from two meaningful units (plus the morphological tag), {\tt <+NN><Neut><Nom><Sg><NA> Wertpapier<NN>@@ 
Konto}. In the post-processing step, the inflected word as well as the accompanying demonstrative can
be generated based on the respective morphological tags.

In the baseline system, the word is built from the more complicated (and non-meaningful) sequence 
{\it Wertpapier@@ kon@@ to}. While the form {\it Wertpapierkonto} itself is correct, the baseline failed
to generate the correct article ({\it dieser} instead of {\it dieses}). 
The segmentation system has incorrectly formed a plural corresponding
to the English {\it accounts}, instead of the correct singular corresponding to {\it account}.

While it is difficult to give a definite explanation, it is possible that the structure of the 
subsequent noun plays a role.
Since the morphological features of a compound are determined by its head noun, in this case {\it konto}
({\it account}), the baseline has struggled here as the word got segmented through BPE. 
Such unfortunate splittings might be less problematic in simpler contexts (e.g. {\it dieses Kon@ to}: 
{\it this account}), but the addition of a modifier between head and article increases the difficulty for 
the system.

The example above, as well as the example in table~\ref{table:example-himl-transformer}, aim at illustrating 
that improvements with the lemma-tag and segmentation strategies are obtained rather indirectly, and are 
difficult to capture and explain.
We assume that the reduction of word forms and the more consistent representation, in combination with
the morphologically annotated tags, play an important role in generating correct word forms given the
context, and consequently in the overall performance of the system.


\subsection{Handling out-of-domain data}
\label{subsection:out-of-domain}
Table~\ref{table:example-himl} shows translations obtained with the baseline and lemma-tag 
Nematus systems, as well as the transformer segmentation system.
In this example, the lemma-tag system managed to produce an acceptable translation for {\it
ultrasound treatment}, whereas the other systems translated {\it sound} as {\it reasonable}
or {\it healthy}.
Furthermore, the translation of the second part of the sentence makes it possible for a reader to guess 
the intended meaning, while the baseline translation contains unrelated words ({\it skin colour} instead
of {\it skin}, and {\it lecture} instead of {\it fracture}).
Similarly, the translation {\it Bruchst\"atte} ({\it rupture place}) in the segmentation system is semantically 
closer to {\it fracture} than the baseline translation, and thus better to understand.


\begin{table}
  \caption{Example: outputs of the baseline Nematus system (B) vs.\ the
  lemma-tag Nematus system (LT) vs.\ the segmentation Transformer system (S),
  all trained on the large corpus, for a sentence from the HimL test set.}
\begin{minipage}{\textwidth}
{\small
\begin{tabular}{p{0.5cm}l}
\hline
\hline
{\bf B} & In der Regel bedeutet die {\bf ultragesunde Behandlung}, dass ein spezielles Ger\"at, \\
        & das mit der Hautfarbe in Ber\"uhrung kommt, t\"aglich rund 20 Minuten lang an \\
        &  der Vorlesung liegt. \\[1ex]
        & \it{typically, the ultra-healthy treatment means that a special device} \\
        & \it{that touches the skin colour abuts the lecture daily for roughly 20 minutes} \\
\hline
{\bf LT} & In der Regel geht es bei der {\bf ultraschall-Behandlung} darum, ein spezielles Ger\"at \\
         & mit der Haut in Kontakt zu bringen, das die Fraktur f\"ur etwa 20 Minuten \\
         & t\"aglich \"uberzieht. \\[1ex]
         & \it{typically, the ultrasound treatment means to bring a special device in contact}\\
         & \it{with the skin, that covers the fracture for around 20 minutes daily}\\
\hline
{\bf S}  & Normalerweise besteht eine {\bf ultravern\"unftige Behandlung} darin, t\"aglich etwa  \\
         & 20 Minuten lang eine Sondervorrichtung in Kontakt mit der Haut zu stellen, die \\
         & die Bruchst\"atte \"uberschwemmt. \\[1ex]
         & \it{usually, an ultra-reasonable treatment consists in putting for around 20 minutes} \\
         & \it{daily a gadget in contact with the skin, which swamps the rupture place} \\
\hline
{\bf REF} & In der Regel umfasst die {\bf Ultraschallbehandlung} die Unterbringung einer \\ 
          &  besonderen Vorrichtung in den Kontakt mit der Haut \"uber der Frakturstelle \\
          & f\"ur etwa 20 Minuten t\"aglich. \\
\hline
{\bf SRC} & Typically , {\bf ultrasound treatment} involves placing a special device in contact \\
          & with the skin overlying the fracture site for around 20 minutes on a daily basis. \\ 
\hline
\end{tabular}
  }
\end{minipage}
\label{table:example-himl}
\end{table}


\section{Other language pair: English--Czech}
\label{sec:en-cz}

To verify that the two strategies also work for another language pair, we re-implemented them
to translate from English into Czech, which has a very rich morphology with regard to {\it case}. 
\cite{huck-EtAl:2017:EACLshort} have previously highlighted how inflection
often leads to out-of-vocabulary problems in English--Czech and have proposed
a solution for phrase-based statistical machine translation.

\paragraph{Data and setup.}
For this experiment, we used a small and a medium training corpus, with data from the WMT
translation shared task.
The small training corpus (203.570 parallel sentences) is obtained from the news-commentary corpus, 
the medium training corpus is a concatenation of news-commentary, common-crawl and Europarl,
resulting in 936.046 parallel sentences.
As development and test sets, we used newstest2015 and newstest2016. 
We applied the same general pre-processing steps (tokenization, true-casing and filtering for length)
as in the English--German setup.
Furthermore, the hyperparameter settings for the translation experiments with the {\it Transformer} model
are the same.

\paragraph{Data pre-processing: Lemma-tag strategy.} 
The lemma-tag strategy for translating into Czech relies on the output of the morphological tagger
{\it Morphodita} \citep{morphodita}. In a first step, the data is tagged and lemmatized. Then, the
surface forms in the training data for the lemma-tag system are replaced by the respective pair of
lemma and morphological tag.
To generate inflected forms for the translation output, {\it Morphodita} is given the lemma-tag
pair and outputs an inflected form. In some instances, {\it Morphodita} generates several forms
for one lemma-tag pair. If this is the case, a word frequency list is used to select the most 
frequent form, in order to reflect preferences for e.g. orthographic variations.

In contrast to the English--German system, the English--Czech system is much simpler as it
relies on only one analysis tool ({\it Morphodita}) instead of combining the output of two 
tools (the parser BitPar and the morphological resource SMOR). 

The English--Czech system also shows that, given an adequate analysis tool, an implementation for
another language pair is rather straightforward.

\paragraph{Data pre-processing: Segmentation strategy.} 
Linguistically informed word segmentation of Czech follows the very same basic
idea as previously adopted for German (Section~\ref{sec:segmentation-pipeline}),
but for Czech we omit the compound splitter component. We cascade a Czech
stemmer-based suffix splitter with BPE segmentation applied on top of the
suffix-split data. 
An out-of-the-box Python
implementation\footnote{\url{http://research.variancia.com/czech_stemmer/}} of
the Czech stemming approach by \cite{DOLAMIC2009714} is modified for our
purposes, i.e., we alter the stemmer's code to not remove morphological
suffixes but rather detach them from the stem and write them out as separate
tokens. In order for the stemmer to purely act as a segmenter, we furthermore
deactivate its integrated functionality for palatalization of stems. 
We evaluate both a light variant and an aggressive variant of linguistic suffix
splitting with the Czech stemmer. The light variant is limited to treating
Czech {\it case} and {\it possessive}. The aggressive variant deals with {\it case}, {\it possessive},
{\it comparative}, {\it diminutive}, {\it augmentative}, and {\it derivational suffixes}.


\begin{table}
\caption{Results with Transformer for the English--Czech LemmaTag and Segmentation approaches in comparison 
  to a system trained on surface forms (Baseline) in case-sensitive \BLEU. Significant improvements over the baseline (p-value=0.05) are marked with~*.}
{\small
\begin{minipage}{\textwidth}
\setlength\tabcolsep{0.5em}
\begin{tabular} {llll}
  \hline
  \hline
  \multicolumn{3}{c}{\bf English--Czech In-domain Translation} \\
  \hline
           & {\bf Small }  & {\bf Medium} \\
  \hline  
    {\bf Transformer BPE Baseline}              & \rdm{10.86} & \rdm{20.12} \\
    {\bf Transformer LemmaTag}                  & \rdm{14.46}\,* & \rdm{20.35} \\
    {\bf Transformer Segmentation (light)}      & \rdm{13.11}\,* & \rdm{20.39} \\
    {\bf Transformer Segmentation (aggressive)} & \rdm{13.17}\,* & \rdm{20.03} \\
  \hline 
\end{tabular}
\end{minipage}
}
\label{table:results-transformer-encz}
\end{table}


\paragraph{Results and discussion.}
Table~\ref{table:results-transformer-encz} shows the results for experiments with small and medium-sized
training data. While the segmentation system and the lemma-tag system improve a lot when using only a small 
training corpus, there is considerably less improvement when using more training data. 
We assume that this is due to Czech's morphological richness that is not nearly sufficiently represented in 
the baseline system in the small setting, which leads to the lemma-tag representation and the segmentation 
strategy having a significant impact here.


\section{Related work}
\label{sec:rel-work}
The modeling of morphology in machine translation has been presented and analyzed in many variants and
settings, both for SMT and NMT scenarios.

For the generation of target-side morphology, many SMT systems made use of the so-called {\it 2-step approach},
in which a translation system first translates into an intermediate representation of the target-side data,
and then applies a prediction step. In this second step, a model separate from the translation system
can predict either directly a surface form, e.g. \citet{toutanova:acl2008}, or grammatical features to be
used for the generation of surface forms, for example \citet{bojar:wmt2010} and \citet{fraser:eacl2012}.
An alternative to the 2-step approach has been presented by \citet{chahuneau:emnlp2013} who
integrated synthetic phrases into the phrase-table of a phrase-based SMT system.

For an NMT setting, \citet{tamchyna17:wmt} presented a strategy to generate inflected forms from pairs of
lemmas and tags for the language pairs English--German and English--Czech; the lemma-tag strategy described 
in this work is based on their system setup.
The lemma-tag strategy discussed in this work --
and
in \citet{tamchyna17:wmt} -- is conceptually related to 
the two-step approach in that the translation system operates on an abstract intermediate representation
followed by the generation of inflected forms. 
However, with the lemmas and full morphological tags being output as a sequence by the translation system, 
there is no separate prediction step required. The second step thus consists only of a deterministic mapping 
between pairs of lemmas and tags to the corresponding inflected form. Note that an interesting alternative idea to this
is to use character decoding guided by morphological information \citep{passban18:naacl}.

The work of \citet{nadejde-etal-2017-predicting} showed that interleaving words and CCG super tags in the training 
data on the source and target side improves the translation quality.
While we do not use CCG tags in our annotation, the morphological tags still provide shallow syntactic information
(though only on the target side), and might even have an effect that goes beyond local agreement.

In the research area of word segmentation, several alternatives to BPE
\citep{sennrich-bpe:acl2016} have been proposed. 
Google is using an in-house sub-word unit segmentation algorithm
\citep{6289079} within their NMT systems \citep{wu-google} to split words into
``word\-pieces''. The Google word\-piece model relies on a language model
component, which BPE does not. 
BPE has gained more widespread acceptance in MT research and is in common use. 
\citet{ataman:eamt2017} and \citet{ataman:acl2018} considered both
supervised and unsupervised splitting of agglutinative morphemes in
Turkish, where Turkish is the source language. 
Their setup addresses the issues arising from translating out of a morphologically
rich language, whereas our work focuses on translating into a language with rich morphology.  
Furthermore, an important linguistic difference here is that
Turkish is an agglutinative language, while German has fusional
inflection and very productive compounding and Czech has highly
productive fusional inflection.

\citet{Pinnis2017} studied more effective segmentation (e.g., by
modifying the BPE algorithm) and \citet{banerjee18:sclem} combined
linguistic segmentation with BPE.

Table~\ref{table:previouswork} gives an overview of the gains in BLEU between the
baseline and the respective best system for strategies to translate into morphologically complex
languages, as discussed above. The systems referred to in this table use RNN/Nematus architectures,
with gains for the language pair EN$\rightarrow$DE lying between 0.47 and 0.7, which is in the same
order of magnitude than our improvements for the RNN system variants, even though it has to be noted
that our training data size is smaller.

A difficult question concerns a more in-depth evaluation of our systems' ability
to handle morphological generalization.
Challenge sets are a crucial area of interest here. We were initially
very interested in carrying out challenge set evaluation using the
challenge sets of \citet{sennrich17:eacl} and \citet{burlot18:wmt},
which capture, e.g., agreement problems, and are very useful for
evaluating BPE. However, it is not clear what the tag representation
in lemma-tag should look like when using challenge sets. If we use the
gold standard tags, then it is very easy for lemma-tag to choose the
correct sentence, but if we use incorrect tags, then lemma-tag cannot
possibly make the correct choice. In practice, the lemma-tag system
does not output incoherent sequences of tags, so it is not clear that
this type of evaluation captures useful facts about lemma-tag. This
is unfortunate, because challenge set evaluation is a clean way of
trying to capture morphological generalization in other contexts.

A test suite analysis focusing on the morphological competence of NMT system variants
can be found in \citet{amrhein-sennrich-2021-suitable-subword}.
They evaluate segmentation strategies on different types of morphological phenomena in
a controlled, semi-synthetic setting. To assess how well NMT systems trained on different
subword and character-level representations can handle various non-concatenative morphological
phenomena such as reduplication or vowel harmony, they insert artificial morphemes to
mimic these phenomena into an German--English translation setting.
The unique morphemes in an otherwise standard data representation allow
to isolate the respective morphological phenomena, and thus to derive which segmentation
strategy is best-suited for a particular morphological problem.
While the analysis in \citet{amrhein-sennrich-2021-suitable-subword} has a different focus
than our main evaluation interest -- measuring the modeling of specific non-concatenative
phenomena vs. capturing morphological generalization as in our work, the method of
isolating particular phenomena is promising and we are interested in taking it further
to evaluate a larger range of morphological issues.


\begin{table}
\small{
\begin{tabular}{ llllr }
\hline
\hline  
    {\bf Work}         & {\bf Modeling} & {\bf System}     & {\bf Lang.}     & {\bf BLEU } \\
                       &                &                  &                 & {\bf gain~~} \\
\hline
 \cite{passban18:naacl} & split target-side  & RNN  & EN$\rightarrow$DE &   0.47 \\   
 & agglutinative morphology          &           &  EN$\rightarrow$RU   & 0.61\\
 & (character-level decoder)          &           & EN$\rightarrow$TR    & 0.69 \\
 \hline
 \cite{nadejde-etal-2017-predicting} & target-side syntax (CCG) & Nematus & DE$\rightarrow$EN & 1.0$^{S}$ \\   
                        & source-/target-side syntax     &         & DE$\rightarrow$EN       & 1.1$^{S}$ \\
                        & source-side syntax (CCG)       &         & EN$\rightarrow$DE & 0.7$^{E}$ \\
                        &  source-side syntax (CCG)       &         & EN$\rightarrow$RO & 0.5$^{E}$ \\
 \hline
 \cite{Pinnis2017}     & morph. guided segmentation       & Nematus & EN$\rightarrow$LV & 0.46$^{*}$ \\   
                       &                                  &         &EN$\rightarrow$LV & 0.71$^{x}$ \\    
 \hline
Banerjee and Bhat- & morph. guided segmentation     & Nematus & BN$\rightarrow$HI & 1.12 \\
tacharyya (2018)       &                     &         & EN$\rightarrow$HI & 0.65 \\
                        &                           &  & EN$\rightarrow$BN & 0.68 \\
\hline
\hline
LemmaTag      & inflection generation       &   Transf. & EN$\rightarrow$DE & -0.2$^{*}$ \\  
Segmentation  & linguistic segmentation     &   Transf. & EN$\rightarrow$DE & 0.0$^{*}$ \\
LemmaTag      & inflection generation       &   Transf. & EN$\rightarrow$DE & 0.7$^{x}$ \\   
Segmentation  & linguistic segmentation     &   Transf. & EN$\rightarrow$DE & 0.4$^{x}$ \\
\hline
LemmaTag      &inflection generation      &   Nematus & EN$\rightarrow$DE & 0.8$^{*}$ \\ 
Segmentation  & linguistic segmentation   &   Nematus & EN$\rightarrow$DE & 0.3$^{*}$ \\
LemmaTag      &inflection generation      &   Nematus & EN$\rightarrow$DE & 0.6$^{x}$ \\ 
Segmentation  & linguistic segmentation   &   Nematus & EN$\rightarrow$DE & 0.5$^{x}$ \\
\end{tabular}
}
\caption{Comparison of our system in terms of BLEU gains with
other
systems for dealing with target-side morphological richness.
For the EN$\rightarrow$DE experiments, \cite{nadejde-etal-2017-predicting} and \cite{passban18:naacl} use $\sim$4.5M training sentences, compared to our $\sim$2M sentences.
~~~~~~~~~~~~~~~~~~~~~~~~~~~~~~~~~~   $^{S}$:single system ~$^{E}$:ensemble model; ~ $^{*}$:news domain ~ $^{x}$:other domain}

\label{table:previouswork}
\end{table}


\section{Conclusion}
\label{sec:conclusion}

In this paper we compared two linguistically informed approaches to dealing with target-side
morphology. The knowledge-rich approach represents words as a combination of a lemma and a
rich POS tag, which is a sufficient representation for generating the final surface forms.
The knowledge-poor approach uses a simple linguistic segmentation strategy based on the
information in a stemmer.

Our experiments show that the Transformer does not benefit as much from these approaches
as the previous shallow-RNN systems did when translating in-domain test sets. However,
for the out-of-domain scenario, where linguistic generalization is more challenging, both
approaches produced large gains.

The knowledge-rich approach depends strongly on the performance of
statistical disambiguation to produce the lemma-tag representation of
the training corpus, and also on the coverage of the inflectional
resource for producing the final translation output. Improvements in
these resources could result in better translation performance. But
the requirement to have such resources is an important aspect of this
approach. If such resources are available for a particular language
pair, then it is an interesting approach to try.

The knowledge-poor approach depends only on having the sort of
information that stemmers typically have access to, with just a small
amount of hand-engineering necessary. Improved stemmers may slightly
improve results, but will probably not drastically change the
translation performance. This approach (and related approaches with
stronger modeling of derivational morphology) should be available for
many morphologically rich languages, and should certainly be tried
for many language pairs for which only BPE-based systems have been
built so far.

In summary, our study has shown that in comparison with the almost-linguistic-knowledge-free
approach of BPE, which only requires substring frequencies, we can achieve important linguistic
generalization through the use of knowledge-rich and knowledge-poor approaches, and that this effect
is particularly strong
in the case of translating out-of-domain.

\section*{Remarks}
This paper was originally accepted for publication in a journal special edition in early 2020.
However, due to the special edition being canceled entirely, this paper has remained unpublished. 
After much consideration, we decided to publish this study on arXiv.

Since writing this paper, we extended the lemma-tag approach to model word formation with a particular focus on handling non-concatenative morphological processes
\citep{weller-di-marco-fraser-2020-modeling},
where we show that linguistic segmentation combined with morpho-syntactic information on both
the source and the target side leads to improvements.

\section*{Acknowledgment}

This research publication was partially funded by LMU Munich's
Institutional Strategy LMUexcellent within the framework of the German
Excellence Initiative.
This project has received funding from the European Research Council
(ERC) under the European Union's Horizon 2020 research and innovation
programme (grant agreement \ftextnumero~640550). This work was
supported by the Dutch Organization for Scientific Research (NWO) VICI
Grant nr.\ 277-89-002.

\newcommand{\newblock}{}
\bibliographystyle{apalike}
\bibliography{bib_file.bib}

\label{lastpage}

\end{document}